\documentclass{article}

\usepackage[final]{neurips_2019}
\usepackage{natbib}

\usepackage [utf8]{inputenc} 
\usepackage [T1]{fontenc}    
\usepackage{hyperref}       
\usepackage{url}            
\usepackage{booktabs}       
\usepackage{amsfonts}       
\usepackage{nicefrac}       
\usepackage{microtype}      

\usepackage{hyperref}
\usepackage{url}
\usepackage{caption}
\usepackage{graphicx}
\usepackage{subfigure}
\usepackage{array}
\usepackage{todonotes}
\usepackage{amsmath,amsfonts,bm}

\DeclareMathOperator*{\E}{\mathbb{E}}   
\DeclareMathOperator*{\V}{\mathbb{V}}   

\DeclareMathOperator*{\Loss}{\mathcal{L}}

\title{Efficient and Robust Reinforcement Learning with Uncertainty-based Value Expansion}

\author{%
  Bo~Zhou \qquad Hongsheng Zeng \qquad Fan Wang  \qquad Yunxiang Li  \qquad Hao Tian\\
  Baidu, Shenzhen, China\\
  \texttt{\{zhoubo01, zenghongsheng, wang.fan, liyunxiang06, tianhao\}@baidu.com} \\
}

\begin{document}

\maketitle

\begin{abstract}
By integrating dynamics models into model-free reinforcement learning (RL) methods, model-based value expansion (MVE) algorithms have shown a significant advantage in sample efficiency as well as value estimation. However, these methods suffer from higher function approximation errors than model-free methods in stochastic environments due to a lack of modeling the environmental randomness. As a result, their performance lags behind the best model-free algorithms in some challenging scenarios. In this paper, we propose a novel Hybrid-RL method that builds on MVE, namely the Risk Averse Value Expansion (RAVE). With imaginative rollouts generated by an ensemble of probabilistic dynamics models, we further introduce the aversion of risks by seeking the lower confidence bound of the estimation. Experiments on a range of challenging environments show that by modeling the uncertainty completely, RAVE substantially enhances the robustness of previous model-based methods, and yields state-of-the-art performance. With this technique, our solution gets the first place in NeurIPS 2019: Learn to Move.
\end{abstract}

\section{Introduction}
In contrast to the tremendous progress made by model-free reinforcement learning algorithms in the domain of games \citep{mnih2015human,silver2017mastering,alphastarblog}, and biomechanical control, poor sample efficiency has risen as a great challenge to RL, especially when interacting with the real world.  A promising direction is to integrate the dynamics model to enhance the sample efficiency of the learning process \citep{sutton1991dyna, Calandra2016,kalweit2017uncertainty,oh2017value,racaniere2017imagination}. However, classic model-based reinforcement learning (MBRL) methods often lag behind the model-free methods (MFRL) asymptotically, especially in stochastic environments where the dynamics models are difficult to learn. The hybrid combination of MFRL and MBRL (Hybrid-RL for short) also attracted attention. A lot of efforts has been devoted to this field, including the Dyna algorithm \citep{sutton1991dyna}, model-based value expansion \citep{feinberg2018model}, I2A \citep{weber2017imagination}, etc. 

The robustness of the learned policy is another concern in RL. In MFRL, off-policy RL typically suffers from this problem and the performance drops suddenly \cite{duan2016benchmarking}. A promising solution is to avoid risk decisions. Risk-sensitive MFRL not only maximizes the expected return, but also tries to reduce those catastrophic outcomes \citep{garcia2015comprehensive, dabney2018implicit,pan2019risk}. For MBRL and Hybrid-RL, without modeling the uncertainty in the environment (especially for continuous states and actions), it often leads to higher function approximation errors and poorer performances. It is proposed that complete modeling of uncertainty in transition can obviously improve the performance \citep{chua2018deep}, however, reducing risks in MBRL and Hybrid-RL has not been sufficiently studied yet.

To achieve sample efficiency and robustness at the same time, we propose a new Hybrid-RL method more capable of solving stochastic and risky environments. The proposed method, namely Risk Averse Value Expansion (RAVE), is an extension of the model-based value expansion (MVE) \citep{feinberg2018model} and stochastic ensemble value expansion (STEVE) \citep{buckman2018sample}. We analyse the higher approximation issue of model-based methods in stochastic environments. Based on the analysis, we borrow ideas from the uncertainty modeling \cite{chua2018deep} and risk averse reinforcement learning. The probabilistic ensemble environment model captures not only the variance in estimation (also called epistemic uncertainty), but also stochastic transition nature of the environment (also called aleatoric uncertainty). Utilizing the ensemble of estimations, we further adopt a dynamic confidence lower bound of the target value function to make the policy more risk-sensitive. We compare RAVE with prior MFRL and Hybrid-RL baselines, showing that RAVE not only yields SOTA expected performance, but also facilitates the robustness of the policy. With this technique, our solution gets the first place in NeurIPS 2019 "Learn to Move" challenge, with a gap of 144 points over the second place\footnote{https://www.aicrowd.com/challenges/neurips-2019-learn-to-move-walk-around/leaderboards.}.

\section{Related Works}

The \textbf{model-based value expansion} (MVE) \citep{feinberg2018model} is a Hybrid-RL algorithm. Unlike typical MFRL such as DQN that uses only 1 step bootstrapping, MVE uses the imagination rollouts to predict the target value. Though the results are promising, they rely on the task-specfic tuning of the rollout length. Following this thread, stochastic ensemble value expansion (STEVE) \citep{buckman2018sample} adopts an interpolation of value expansion of different horizon lengths. The accuracy of the expansion is estimated through the ensemble of environment models as well as value functions. Ensemble of environment models also models the uncertainty to some extent, however, ensemble of deterministic model captures mainly epistemic uncertainty instead of stochastic transitions \citep{chua2018deep}.

The \textbf{uncertainty} is typically divided into three classes \citep{Geman92Neural}: the \emph{noise} exists in the objective environments, e.g., the stochastic transitions, which is also called aleatoric uncertainty \citep{chua2018deep}. The \emph{model bias} is the error produced by the limited expressive power of the approximating functions, which is measured by the expectation of ground truth and the prediction of the model, in case that infinite training data is provided. The \emph{variance} is the uncertainty brought by insufficient training data, which is also called epistemic uncertainty. \cite{dabney2018distributional} discuss the epistemic and aleatoric uncertainty in their work and focus on the latter one to improve the distributional RL. Recent work suggests that ensemble of probabilistic model (PE) is considered as more thorough modeling of uncertainty \citep{chua2018deep}, while simply aggregate deterministic model captures only epistemic uncertainty. The aleatoric uncertainty is more related to the stochastic transition, and the epistemic uncertainty is usually of interest to many works in terms of exploitation \& exploration \citep{pathak2017curiosity,schmidhuber2010formal,oudeyer2009intrinsic}. Other works adopt ensemble of deterministic value function for exploration \citep{osband2016deep, buckman2018sample}. 

\textbf{Risks} in RL typically refer to the inherent uncertainty of the environment and the fact that policy may perform poorly in some cases \citep{garcia2015comprehensive}. Risk sensitive learning requires not only maximization of expected rewards, but also lower variances and risks in performance. Toward this object, some works adopt the variance of the return \citep{sato2001td,pan2019risk,reddy2019risk}, or the worst-case outcome \citep{heger1994consideration,gaskett2003reinforcement} in either policy learning \citep{pan2019risk, reddy2019risk}, exploration \citep{smirnova2019distributionally}, or distributional value estimates \citep{dabney2018implicit}. An interesting issue in risk reduction is that reduction of risks is typically found to be conflicting with exploration and exploitation that try to maximize the reward in the long run. Authors in  \citep{pan2019risk} introduce two adversarial agents (risk aversion and long-term reward seeking) that act in combination to solve  to trade-off between risk-sensitive and risk-seeking (exploration) in RL. In this paper, we propose a dynamic confidence bound for this purpose.

A number of prior works have studied function approximation error that leads to overestimation and sub-optimal solution in MFRL. Double DQN \citep{van2016deep} improves over DQN through disentangling the target value function and the target policy that pursues maximum value. The authors of TD3 \citep{Fujimoto2018Addressing} suggest that systematic overestimation of value function also exists in actor-critic MFRL. They use an ensemble of two value functions, with the minimum estimate being used as the target value. Selecting the lower value estimation is similar to using uncertainty or lower confidence bound which is adopted by the other risk sensitive methods \citep{pan2019risk}, though they works have different motivations.

\section{Preliminaries}

\subsection{Actor-Critic Model-free Reinforcement Learning}

The Markov Decision Processes(MDP) is used to describe the process of an agent interacting with the environment. The agent selects the action $a_t \in \mathcal{A}$ at each time step $t$. After executing the action, it receives a new observation $s_{t+1} \in \mathcal{S}$ and a feedback $r_t \in \mathbb{R}$ from the environment. As we focus mainly on the environments of continuous action, we denote the \emph{policy} that the agent uses to perform actions as $a_t = \pi(s_t)$. As the interaction process continues, the agent generates a trajectory $\tau = (s_0, a_0, r_0, s_1, a_1, r_1, ...)$ following the policy $\pi$. For finite horizon MDP, we use the indicator $d: \mathcal{S} \rightarrow \{0,1\}$ to mark whether the episode is terminated. The objective of RL is to find the optimal policy $\pi^*$ to maximize the expected discounted sum of rewards along the trajectory. The value performing the action $a$ with the policy $\pi$ at the state $s$ is defined by $Q^{\pi}(s, a) = \E_{a \sim \pi} \{\sum_{t=0}^{\infty}\gamma^{t}r_t|s_0=s, a_0=a \}$, where $0 < \gamma < 1$ is the discount factor. The Q-value function can be updated with Bellman equation, by minimizing the Temporal  Difference(TD) error:
\begin{equation}
\begin{split}
&\Loss(Q) = \E_{\tau} \left[ \sum_t(r_t + \gamma \cdot \hat{Q}(s_{t+1}, a) - Q(s_{t}, a_{t})) ^2 \right]\\
\label{TD-Learning}
\end{split}
\end{equation}
The target Q-values are estimated by a target network $\hat{Q}_\phi'$, where $\phi'$ is a delayed copy of the parametric function approximator $Q_\phi$ \cite{lillicrap2015continuous}.

To optimize the deterministic policy function in a continuous action space, deep deterministic policy gradient(DDPG) maximizes the value function (or minimizes the negative value function) under the policy $\pi$:
\begin{equation}
\Loss(\pi) = -\E_{\tau}[\sum_t Q(s_t, \pi(s_t))]
\label{PolicyGradient}.
\end{equation} 

\subsection{Environment Modeling}
To model the environment in continuous space, an environment model is typically composed of three individual mapping functions: $\hat{f}_{r}: \mathcal{S}\times \mathcal{A} \times \mathcal{S} \rightarrow \mathbb{R}$, $\hat{f}_{s}: \mathcal{S}\times \mathcal{A} \rightarrow \mathcal{S}$, and $\hat{f}_{d}: \mathcal{S} \rightarrow \left[0, 1\right]$, which are used to approximate the feedback, next state and probability of the terminal indicator respectively\cite{gu2016continuous,feinberg2018model}. With the environment model, starting from $s_t, a_t$, we can predict the next state and reward by
$\hat{s}_{t+1} = \hat{f}_{s}(s_t,a_t), \hat{r}_{t} = \hat{f}_{r}(s_t, a_t, \hat{s}_{t+1}), \hat{d}_{t+1} = \hat{f}_{d}(\hat{s}_{t+1})$,
and this process might go on to generate a complete \textit{imagined trajectory} of $[s_t, a_t, \hat{r}_{t}, \hat{s}_{t+1}, ...]$.

\subsection{Uncertainty Aware Prediction}

The deterministic model approximates the \emph{expectation} only and cannot capture either aleatoric or epistemic uncertainty. Following the recent work that studies the uncertainty\cite{chua2018deep}, we briefly review different uncertainty modeling techniques. 

\textbf{Probabilistic model} outputs a distribution  (e.g., mean and variance of a Gaussian distribution) instead of an expectation. Taking the reward component of the environment model as an example, the probabilistic model is written as $r \sim \mathcal N (\hat{f}_{r}, \hat{\sigma}_{r}^2)$, and the loss function is the negative log likelihood:
\begin{equation}
\Loss(\hat f_r) = -\E_{\tau}\left [\log {p_{\mathcal{N}} (r_t | \hat{f}_{r} (s_t, a_t, s_{t+1}), \hat{\sigma}_{r}^2 (s_t, a_t, s_{t+1}))}\right].
\label{Probablistic_Loss}
\end{equation}

\textbf{Ensemble of deterministic} (DE) model maintains an ensemble of parameters, which is typically trained with a unique dataset. Given the ensemble of parameters $\bm{\zeta}$=$\{\zeta_{1}, \zeta_{2}, ..., \zeta_{N}\}$, the variance of predicted values $\V \left [\hat{f}_{r, \bm{\zeta}}\right] = \frac{1}{N}\sum_i  (\hat{f}_{r, \zeta_{i}} - \E \left [\hat{f}_{ r, \bm{\zeta}}\right])^2$ measures the prediction uncertainty.

The variance $\hat{\sigma}^2$ in equation (\ref{Probablistic_Loss}) mainly captures the \emph{aleatoric} uncertainty, and the variance $\V$ mainly captures the \emph{epistemic} uncertainty \citep{chua2018deep}.

\textbf{Ensemble of probabilistic models} (PE) keeps track of an collection of distributions  $\{\mathcal N (\hat{f}_{r, \zeta_{i}}, \hat{\sigma}^2_{r, \zeta_{i}})\}, i \in  [1,N]$, which can capture the aleatoric uncertainty as well as epistemic uncertainty.

\subsection{Model-Based Value Expansion}
MVE\cite{feinberg2018model} uses the learned environment model $\hat{f} = (\hat{f}_{s,}, \hat{f}_{r}, \hat{f}_{d})$ and the policy $\pi$ to imagine a trajectory. We define the imaginative trajectory with the rollout horizon H as $\hat{\tau}_{H}$($H \geq 0$). For the imaginative trajectory starting from state $s_t$ and action $a_t$, we can write $\hat{\tau}_H(r_t, s_{t+1}) = (r_t, s_{t+1}, \hat{a}_{t+1}, \hat{r}_{t+1}, \hat{s}_{t+2}, ..., \hat{s}_{t+H+1}, \hat{a}_{t+H+1})$.

MVE defines the target values based on the imaginative trajectory $\hat{H}(r_t, s_{t+1})$ as:

\begin{equation}
\begin{split}
\hat{Q}^{\text{MVE}}_{H}(r_t, s_{t+1}) = r_t + \sum_{t'=t+1}^{t+H}\gamma^{t'-t} d_{t,t'} \hat{r}_{t'}  + \gamma^{H+1} d_{t,t+H+1} \hat{Q}(\hat{s}_{t+H+1}, \hat{a}_{t+H+1})\\
\text{with } d_{t,t'} =  (1 - d(s_{t+1}))\prod_{k=t+2}^{t'}(1 - \hat{f}_{d}(\hat{s}_k)) \qquad\qquad\qquad\nonumber
\end{split}
\label{TD_MVE}
\end{equation}

\subsection{Stochastic Ensemble Value Expansion}
Selecting proper horizon $H$ for value expansion is important to achieve high sample efficiency and asymptotic accuracy at the same time. Though the increase of $H$ brings increasing prediction ability, the asymptotic accuracy is sacrificed due to the increasing reliance on the environment model.

To reduce the difficulty of selecting proper horizons for different environments, STEVE\cite{buckman2018sample} proposes to interpolate the estimated values  $\hat{Q}^{\text{MVE}}_{H}$ of different $H \in [0, H_{max}]$. Modeling the dynamics and the Q function with an ensemble of models, STEVE decides the weight for each rollout step by considering the variance of ensemble predictions. We denote the parameters $\zeta_s, \zeta_r,$ and $\zeta_d$ for the dynamics models $f_s, f_r$ and$f_d$ respectively, and $\phi$ for the Q function. The parameter set for ensemble can be denoted as $\bm{\zeta}_n=\left\{ \zeta_{s,1}, \zeta_{r,1}, \zeta_{d,1}, \phi_1...\zeta_{d,n}, \phi_n\right\}$, where n represents the ensemble size of each function. The target values in STEVE can be expressed as:
\begin{equation}
\begin{split}
&\hat{Q}^{\text{STEVE}}(r_t,s_{t+1}) = \frac{\sum_{H=0}^{H_{max}} \omega_H \E\left[\bm{\hat{Q}^{\text{MVE}}_{H,\bm{\zeta}}}(r_t,s_{t+1})\right]}{\sum_{H=0}^{H_{max}} \omega_H},\\
&\qquad\text{with }\omega_H = \V\left[\bm{\hat{Q}^{\text{MVE}}_{H,\bm{\zeta}}}(r_t, s_{t+1})\right]^{-1},
\end{split}
\label{STEVE}
\end{equation}
where $\bm{\hat{Q}^{\text{MVE}}_{H,\bm{\zeta}}}$ represents the outcomes of ensemble models, with $\E$ and $\V$ representing their mean and variance respectively. More details about the ensemble prediction can be found at the appendix.

\section{Investigation of the approximation error in stochastic environments}\label{sec:demo}
\begin{figure} [!ht]
\centering

\subfigure [Toy Environment]
{
    \begin{minipage} [t]{0.48\textwidth}
    \centering
    \includegraphics [width=6cm,height=1.5cm]{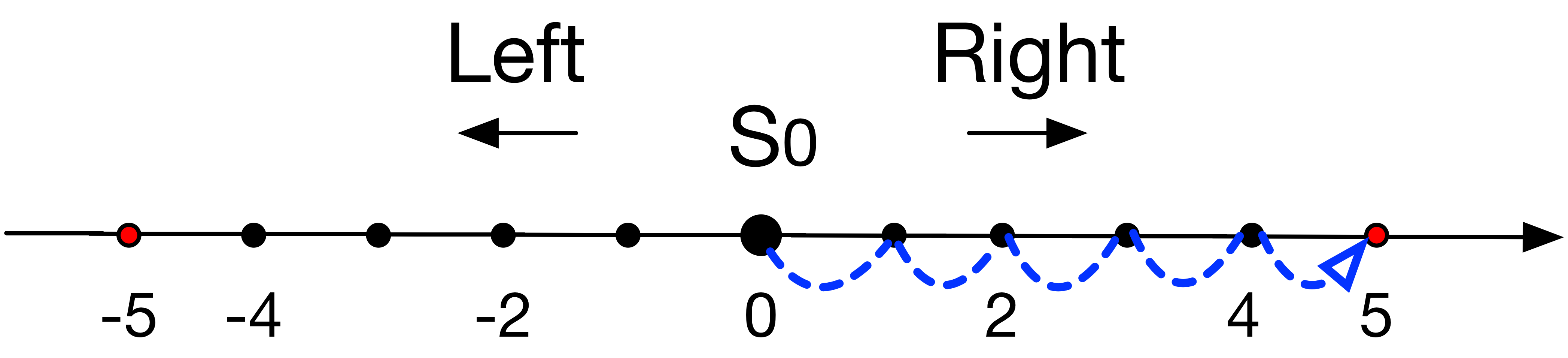}
    \label{fig:ToyEnv}
    \end{minipage}
}

\subfigure [The change of $\hat{Q}_{\phi} (s_0, a_0 = 1)$ with k = 0]
{
    \begin{minipage} [t]{0.48\textwidth}
    \centering
    \includegraphics [width=6cm,height=4cm]{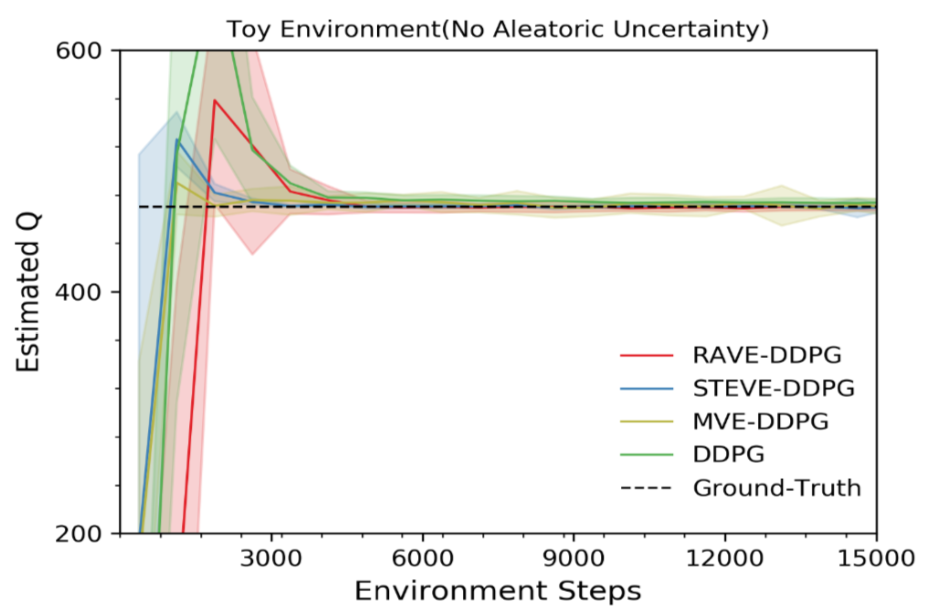}
    \label{fig:ToyEnvLeft}
    \end{minipage}
}
\subfigure [The change of $\hat{Q}_{\phi} (s_0, a_0 = -1)$ with k = 1]
{
    \begin{minipage} [t]{0.48\textwidth}
    \centering
    \includegraphics [width=6cm,height=4cm]{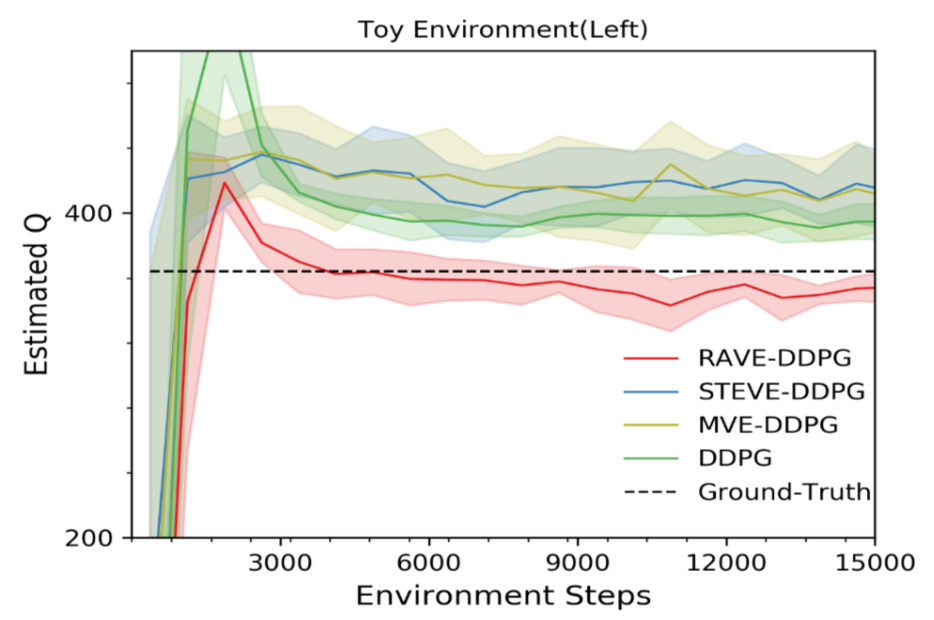}
    \label{fig:ToyEnvQRight}
    \end{minipage}
}
\caption{Curves of estimated Q values in the toy environment at the starting point over the environment steps.Each experiment is run four times.}
\vspace*{-.1in}
\label{fig:ToyExp}
\end{figure}

To thoroughly investigate the impact of incomplete modeling uncertainty on hybri-RL methods, we construct a demonstrative toy environment (fig. \ref{fig:ToyEnv}). The agent starts from $s_0 = 0$, chooses an action $a_t$ from $\mathcal{A} = \left [-1, +1\right]$ at each time step $t$. The transition of the environment is $s_{t+1}=s_t+\frac{a_t}{|a_t|}+k\cdot\mathcal{N} (0,1)$. We compare two different environments:$k = 0$ and $k = 1$, where $k = 0$ represents the deterministic transition, and $k = 1$ represents the stochastic transition. The episode terminates at  ($|s| > 5$ ), where the agent acquires a final reward. The agent gets a constant penalty at each time step to encourage it to reach the terminal state as soon as possible. Note that the deterministic environment actually requires more steps in expectation to reach $|s > 5|$ compared with the stochastic environment, thus the value function at the starting point of $k=1$  (Ground truth = 380+) tends to be lower than that of $k=0$ (Ground truth = 430+). 

We apply DDPG, MVE, STEVE to this environment, and plot the changes of estimate values at the starting point (see fig.~\ref{fig:ToyExp}).

The results show that, in the deterministic environment, the Q-values estimated by all methods converge to the ground-truth asymptotically in such a simple environment. However, after adding the noise, previous MFRL and Hybrid-RL methods show various level of overestimation. The authors of  \citep{feinberg2018model} have claimed that value expansion improves the quality of estimated values, but MVE and STEVE actually give even worse prediction than model-free DDPG in the stochastic environment. A potential explanation is that the overall overestimation comes from the unavoidable imprecision of the estimator \citep{thrun1993issues, Fujimoto2018Addressing}, but Hybrid-RL also suffers from the approximation error of the dynamics model. When using a deterministic environment model, the predictive transition of both environments would be identical, because the deterministic dynamics model tends to approximate the expectation of next states (e.g, $\hat{f}_{s,\zeta_s} (s_t=0, a_t>0) = 1.0, \hat{f}_{s,\zeta_s} (s_t=1.0, a_t>0) = 2.0$). This would result in the same estimated values for $k=0$ and $k=1$ for both value expansion methods, but the ground truth of Q-values are different in these two environments. As a result, the deterministic environment introduces additional approximation error, leading to extra bias of the estimated value.

\section{Methodology}\label{sec:methodology}
\subsection{Risk Averse Value Expansion}
We proposed mainly two improvements based on MVE and STEVE. Firstly, we apply an ensemble of probabilistic models  (PE) to enable the environment model to capture the uncertainty, including aleatoric and epistemic uncertainty. Secondly, inspired by risk sensitive RL, we calculate the confidence lower bound of the target values, using the variance of aleatoric and epistemic uncertainty.


Before introducing RAVE, we start with the Distributional Value Expansion (DVE). Compared with MVE that uses a deterministic environment model and value function, DVE uses a probabilistic environment model, and we independently sample the reward and the next state using the probabilistic environment models:
\begin{equation}
\begin{split}
&\tilde{s}_{t+2} \sim \mathcal N (\hat{f}_{s, \zeta_{s}} (s_{t+1}, \tilde{a}_{t+1}), \hat{\sigma}^2_{s, \zeta_{s}} (s_{t+1}, \tilde{a}_{t+1})), \\
& \tilde{r}_{t+1} \sim \mathcal N (\hat{f}_{r, \zeta_{r}} (s_{t+1}, \tilde{a}_{t+1}, \tilde{s}_{t+2}), \hat{\sigma}^2_{r, \zeta_{r}} (s_{t+1}, \tilde{a}_{t+1}, \tilde{s}_{t+2})), \\
& \tilde{d} (\tilde{s}_{t+2}) \sim \mathcal N (\hat{f}_{d, \zeta_{d}} (\tilde{s}_{t+2}), \hat{\sigma}^2_{d, \zeta_{d}} (\tilde{s}_{t+2})).
\end{split}
\label{DVEExpand}
\end{equation}
We apply the distributional expansion starting from $r_t, s_{t+1}$ to acquire the trajectory $\tilde{\tau}_{\bm{\zeta}, H} (r_t, s_{t+1}) =  (r_t, s_{t+1}, \tilde{a}_{t+1}, \tilde{r}_{t+1}, \tilde{s}_{t+2}, ..., \tilde{s}_{t+H+1}, \tilde{a}_{t+H+1})$, which leads to the definition of DVE:
\begin{equation}
\begin{split}
&\hat{Q}^{\text{DVE}}_{\bm{\zeta}, H} (r_t, s_{t+1}) = r_t + \sum_{t'=t+1}^{t+H}\gamma^{t'-t} d_{t,t'} \tilde{r}_{t'} + \gamma^{H+1} d_{t,t+H+1} \hat{Q} (\tilde{s}_{t+H+1}, \tilde{a}_{t+H+1}),\\
&\quad \text{with } d_{t,t'} =   (1 - d (s_{t+1}))\prod_{k=t+2}^{t'} (1 - \tilde{d} (\tilde{s}_k)).
\end{split}
\label{DVE}
\end{equation}

With an ensemble of models on dynamics and value functions, we can compute multiple estimates based on different combinations of dynamics models and Q functions(details about the combination can be found at the appendix). Then we count the average and the variance on the ensemble of DVE, and by subtracting a certain proportion ($\alpha$) of the standard deviation, we acquire a lower bound of DVE estimation. We call this estimation of value function the $\alpha$-confidence lower bound ($\alpha$-CLB):
\begin{equation}
\hat{Q}^{\alpha-CLB}_{\bm{\zeta}, H} (r_t, s_{t+1}) = \E \left [\hat{Q}^{\text{DVE}}_{\bm{\zeta}, H} (r_t, s_{t+1})\right] - \alpha \sqrt{\V\left [\hat{Q}^{\text{DVE}}_{\bm{\zeta}, H} (r_t, s_{t+1})\right]}.
\label{LBVE_Equation}
\end{equation}
The variance in $\hat{Q}^{\alpha-CLB}$ consists of aleatoric and epistemic uncertainty, and we present the proof of $\hat{Q}^{\alpha-CLB}$ convergence in Appendix using the Bellman backup. Subtraction of variances is a commonly used technique in risk-sensitive RL \citep{sato2000variance, pan2019risk,reddy2019risk}. It tries to suppress the utility of the high-variance trajectories to avoid possible risks. Finally, we define RAVE, which adopts the similar interpolation among different horizon lengths as STEVE based on DVE and CLB:
\begin{equation}
\begin{split}
&\hat{Q}_{target} (r_t, s_{t+1}) \leftarrow \hat{Q}^{\text{RAVE}} (r_t, s_{t+1}) = \frac{\sum_{H=0}^{H_{max}} \omega_H \hat{Q}^{\alpha-\text{CLB}}_H (r_t, s_{t+1})}{\sum_{H=0}^{H_{max}} \omega_H},\\
&\qquad\text{with }\omega_H = \V \left [\hat{Q}^{\text{DVE}}_{\bm{\zeta}, H} (r_t, s_{t+1})\right]^{-1}.
\end{split}
\label{RAVE_Equation}
\end{equation}


\subsection{Adaptive Confidence Bound}

An unsolved problem in RAVE is to select proper $\alpha$. The requirement of risk aversion and exploration is somehow competing: risk aversion seek to minimize the variance, while exploration searches states with higher variance (uncertainty). The agent needs to explore the state space sufficiently, and it should get more risk-sensitive as the model converges. The epistemic uncertainty is a proper indicator that measures how well our model get to know the state space. In MBRL and Hybrid-RL, a common technique to monitor the epistemic uncertainty is evaluating the ability of the learned environment model to predict the consequence of its own actions \citep{pathak2017curiosity}.

We set the confidence bound factor to be related to its current state and action, denoted as $\alpha (s,a)$. We want $\alpha (s,a)$ to be larger when the environment model could perfectly predict the state to get more risk sensitive, and smaller when the prediction is noisy to allow more exploration. We define
\begin{equation}
\alpha (s_t,a_t) = max\{0, \alpha (1.0 - \frac{1}{Z}||\E_{\zeta_s} [\hat{f}_{s, \zeta_s} (s_t, a_t)] - s_{t+1}||^2)\}, 
\label{RAVE_Equation}
\end{equation}
where $Z$ is a scaling factor for the prediction error. With a little abuse of notations, we use $\alpha$ here to represent a constant hyperparameter, and $\alpha (s,a)$ is the factor that is actually used in $\alpha$-CLB. $\alpha (s_t, a_t)$ picks the value near zero at first, and gradually increases to $\alpha$ with the learning process.

\begin{figure*} [ht!]%
       \centering
       \captionsetup{singlelinecheck = false, justification=justified}
       \includegraphics [width=1\textwidth]{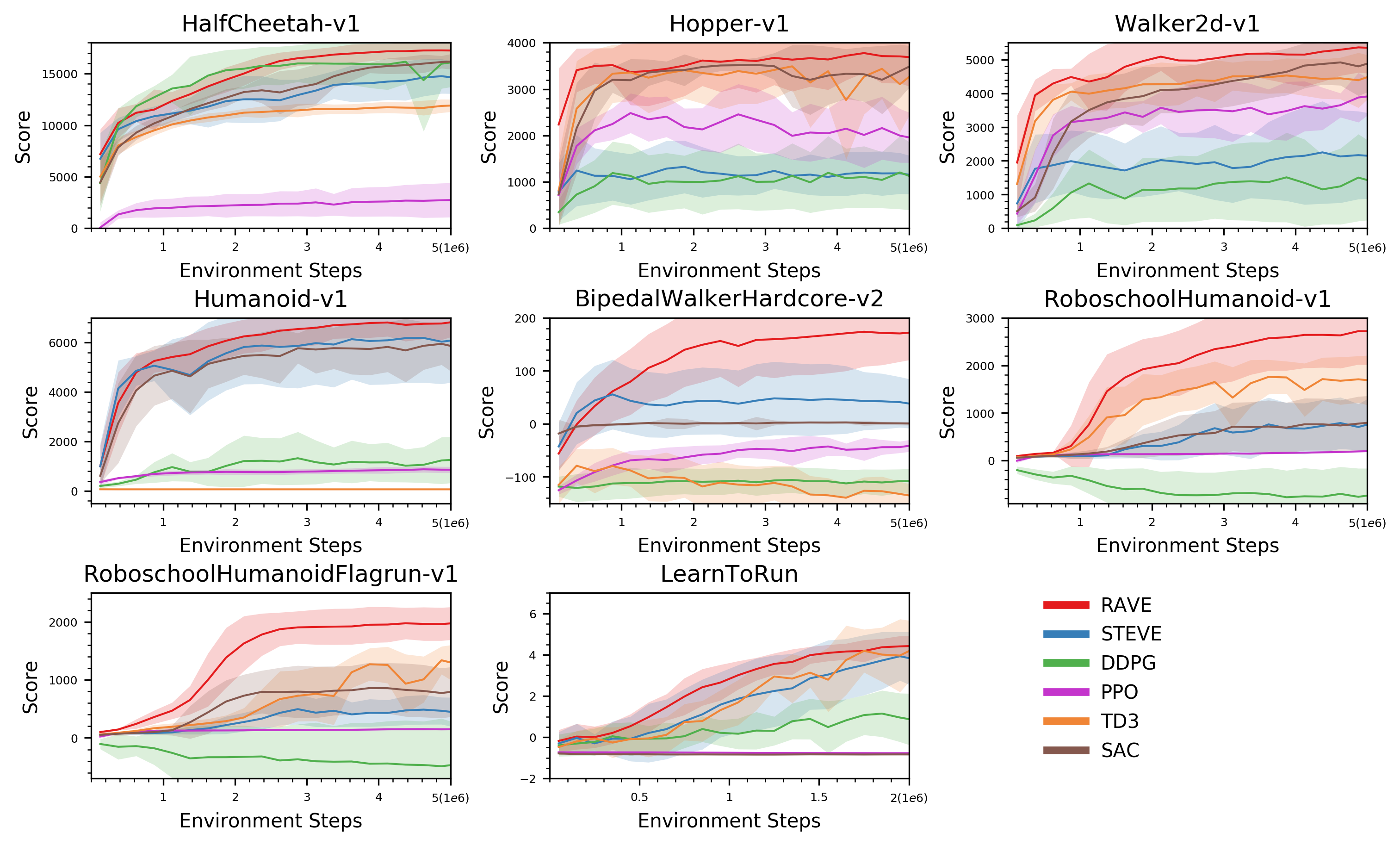}
       \caption{Average return over environment frames in MuJoCo and Roboschool environments. Each experiment is run four times. Experiments on Mujoco and Robo-school environments run 5M steps, and osim-rl experiments run 2M steps due to the time-consuming simulation.}
       \label{fig:Exp}
\vspace*{-.1in}
\end{figure*}

\section{Experiments and Analysis}
We evaluate RAVE on continuous control environments using the MuJoCo physics simulator \citep{todorov2012mujoco}, Roboschool \citep{Klimov2017}, osim-based environment \cite{seth2018opensim}. The baselines includes the model-free DDPG, proximal policy optimization (PPO) \citep{schulman2017proximal}, and STEVE that yields the SOTA Hybrid-RL performance. We also compare RAVE with the SOTA MFRL methods including twin delayed deep deterministic  (TD3) \citep{Fujimoto2018Addressing}, soft actror-critic (SAC) algorithm \citep{haarnoja2018soft}, using the author-provided implementations. To further demonstrate the robustness in complex environments, we also evaluate RAVE on the osim-rl environment with the task: learning to run. For details about the environment setting and implementation\footnote{we present training details of the 1st place solution in the supplementary materials due to the limited space of the paper.}, please refer to the supplementary materials.

\subsection{Experimental Results}
We carried out experiments on nine environments shown in fig.~\ref{fig:Exp}. Among the compared methods, PPO has very poor sample-efficiency, as PPO needs a large batch size to learn stably \citep{haarnoja2018soft}. On Hopper and Walker2d, STEVE lags behind the best MFRL methods (TD3 and SAC), which yield quite good performance in many environments. 
Overall, RAVE performed favorably in most environments in both sample efficiency and asymptotic performance.

\subsection{Analysis}
\begin{figure*} [ht!]%
       \centering
       \captionsetup{singlelinecheck = false, justification=justified}
       \includegraphics [width=1.0\textwidth]{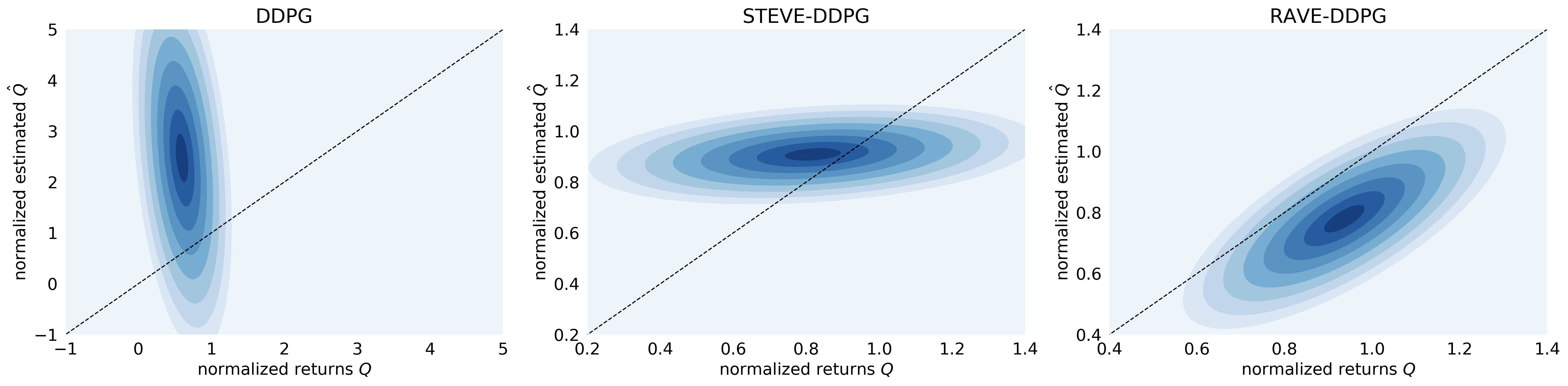}
       \caption{Confidence ellipses of the distribution of estimated values $\hat{Q}$ versus the ground truth in Hopper-v1. The points are extracted from environment steps of 1M, each includes statistics from 10,000 points. The x-axis and y-axis represent the statistical cumulative discounted returns (ground truth) and the predictive Q-values (both are normalized), respectively.}.
       \label{fig:QApproximation}
\vspace*{-.2in}
\end{figure*}

\textbf{Distribution of Value Function Approximation}. While the RAVE estimation on the toy environment has demonstrate its improvement on value estimation, we further investigate whether it predicts value function more precisely in a more complicate environment. We plot the of the predicted values $\hat{Q}$ against the ground truth values of Hopper-v1 in fig.~\ref{fig:QApproximation}. The ground truth value here is calculated by directly adding the rewards of the left trajectory, thus it is more like a Monte Carlo sampling from ground truth distribution, which is quite noisy. To better demonstrate the distribution of points, we draw the confidence ellipses representing the density. The points are extracted from the model at environment steps of 1M. In DDPG and STEVE, the predicted value and ground truth aligned poorly with the ground truth, while RAVE yields better alignments, though a little bit of underestimation.

\begin{figure} [!ht]
\centering

\subfigure [Performance of RAVE variants]
{
    \begin{minipage} [t]{0.48\textwidth}
    \centering
    \includegraphics [width=6cm,height=5cm]{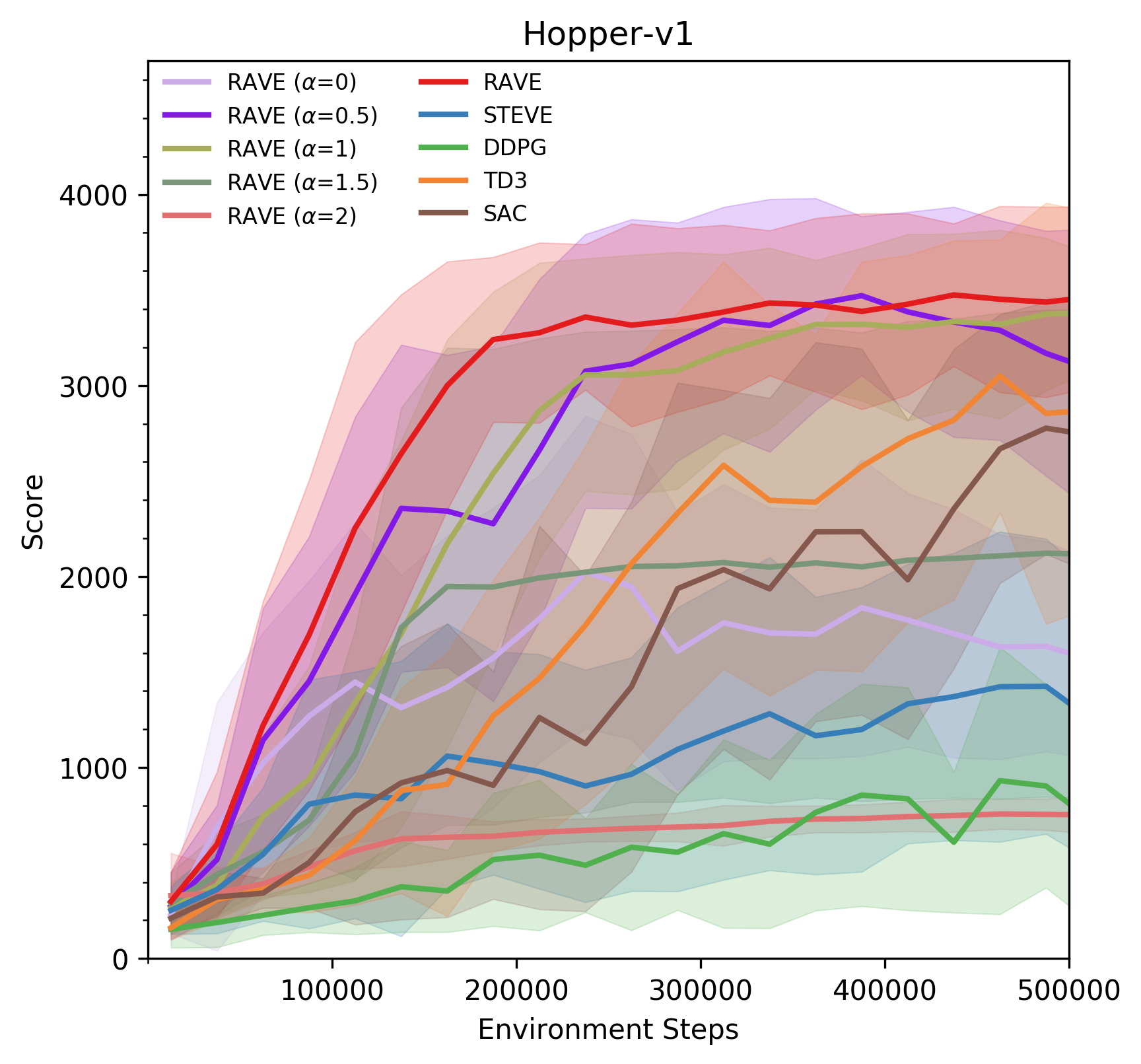}
    \label{fig:ablation}
    \end{minipage}
}
\subfigure [Falling rate]
{
    \begin{minipage} [t]{0.48\textwidth}
    \centering
    \includegraphics [width=6cm,height=5cm]{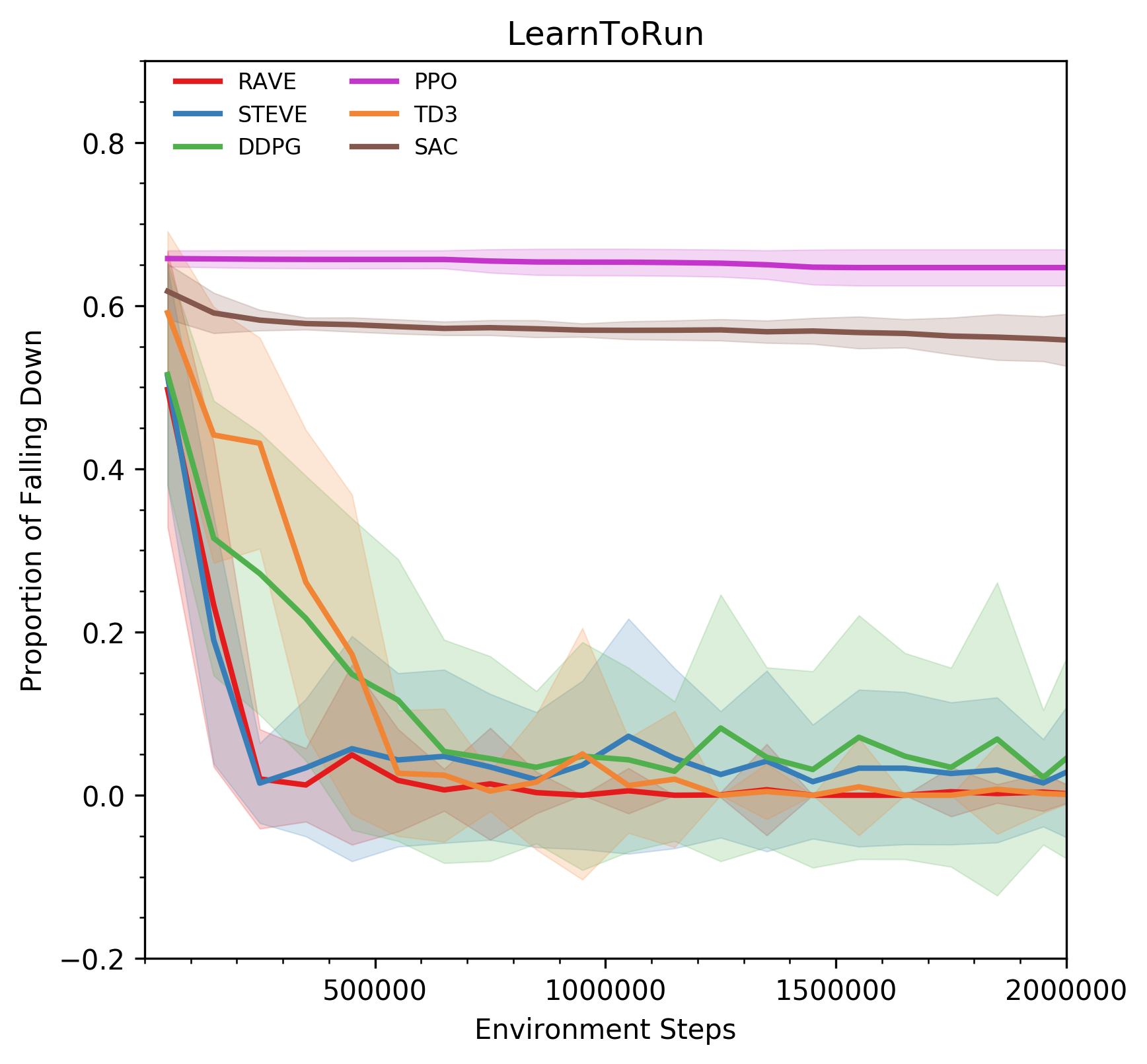}
    \label{fig:FallingRate}
    \end{minipage}
}
\caption{Examining the performance of RAVE variants and robustness of the learned policy.  (a) Performance of RAVE with different $\alpha$-CLB. (b) The falling rate of various algorithms over the 2M environment steps. Each point is evaluated on the outcome of 1000 episodes.} 
\vspace*{-.1in}
\label{fig:albation_exp}
\end{figure}

\textbf{Investigation on dynamic confidence bound}\label{sub:Ablation}.
To study the role played by the $\alpha$-confidence lower bound separately, we further carried out ablative experiments. We compared RAVE ($\alpha$ is a constant), RAVE (dynamic $\alpha$) and other baselines in fig.~\ref{fig:albation_exp}. 

We can see that $\alpha = 0$ already surpasses the performance of STEVE on Hopper, showing that modeling aleatoric uncertainty through PE indeed benefits the performance of value expansion. Larger margin was attained by introducing $\alpha$-CLB. A very large $\alpha$ (such as constant $\alpha = 2.0$, which means lower CLB) can quickly stabilize the performance, but its performance stayed low due to lack of exploration, while a smaller $\alpha$ (constant $\alpha$ = 0.5) generates larger fluctuation in performance. The dynamic adjustment of $\alpha$ facilitates quick rise and stable performance.

\textbf{Analysis on Robustness}. We also evaluated the robustness of RAVE and baselines, by testing the falling rates of the learned agents. We test the perfmance of various algorithms in the LearnToRun task, as it is reported that the agent is prone to fall in this environment \cite{DBLP:journals/corr/abs-1902-02441}. As shown in fig.\ref{fig:FallingRate}, RAVE achieves the lowest falling rate in a short time compared with the baselines. We also found that RAVE played an importance role in the competition, where the future velocity commands depend on the future position. It adds additional bias into Q-value estimation, and in such case, RAVE with risk-aversion Q-values can stabilize the learning policy. In our experiments, DDPG-based policy fell at the probability of 15\%, while the learned agent of RAVE performed more stably, at \textbf{1.3\%}.

\textbf{Computational Complexity}\label{sub:Compute}. A main concern toward RAVE may be its computational complexity. 

For the training stages, the additional training cost of RAVE compared with STEVE rises from the sampling operation. In our experiments, it takes 13.20s for RAVE to finish training 500 batches with a batch size 512, an increase of 24.29\%, compared to STEVE (10.62s)\footnote{The time reported here is tested in 2 P40 Nvidia GPUs with 8 CPUs (2.00GHz), with same number of candidate target values generated for both STEVE and RAVE.}.

For the inference stages, RAVE charges exactly the same computational resources just as the other model-free actor-critic methods as only the learned policy is used for inference.

\section{Conclusion}
In this paper, we raise the problem of incomplete modeling of uncertainty and insufficient robustness in model-based value expansion. To address the issue, we introduce the ensemble of probabilistic models to better approximate the environment, as well as the alpha-Confidence Lower Bound to avoid the opportunistic solution. Based on the ensemble imaginative rollout predictions, we take the lower confidence bound for value estimation to avoid optimistic estimation, which will lead to risky action selections. We also suggest tuning the lower confidence bound dynamically to balance the risk-averse actions and exploration. Experiments on a range of environments demonstrate the superiority of RAVE in both sample efficiency and robustness, compared with state-of-the-art RL methods, including the model-free TD3 algorithm and the Hybrid-RL STEVE algorithm. The RAVE algorithm also provides a plausible model-based and robust solution for the Neurips challenge on the physiologically-based human model. We hope that this algorithm will facilitate the application of reinforcement learning in risky, real-world scenarios.

\clearpage

\bibliography{references}
\bibliographystyle{plainnat}

\clearpage

\appendix

\section{Training and Implementation Details}
\subsection{Neural Network Structure}
We used rectified linear units  (ReLUs) between all hidden layers of all our implemented algorithms. Unless otherwise stated, all the output layers of model have no activation function.

\textbf{RL Models}. We implement model-based algorithms on top of DDPG, with a policy network and a Q-value network. The policy network is a stack of 4 fully-connected (FC) layers. The activation function of the output layer is \textit{tanh} to constrain the output range of the network. The Q-value network takes the concatenation of the state $s_t$ and the action $a_t$ as input, followed by four FC layers.

\textbf{Dynamics Models}. We train three neural networks as the transition function, the reward function and the termination function. We build eight FC layers for the transition approximator, and four FC layers for the other approximators. The distributional models $\mathcal{N} (\hat{f}, \hat{\sigma}^2)$ in RAVE use the similar model structure except that there are two output layers corresponding to the mean and the variances respectively.

\subsection{Parallel Training}
We use distributed training to accelerate our algorithms and the baseline algorithms. Following the implementation of STEVE, we train a GPU learner with multiple actors deployed in a CPU cluster. The actors reload the parameters periodically from the learner, generate trajectories and send the trajectories to the learner (we used 8 CPUs for mujoco and roboschool environments, 128 CPUs for the osim-rl environment). The learner stores them in the replay buffer, and updates the parameters with data randomly sampled from the buffer. For the network communication, we use PARL\footnote{https://github.com/PaddlePaddle/PARL} to transfer data and parameters between the actors and the learner. We have 8 actors generating data, and deploy the learner on the GPUs. For osim tasks, we use 128 actors as its simulation speed is much slower than Mujco.
DDPG uses a GPU, and model-based methods uses two: one for the training of the policy and another for the dynamics model.

\begin{figure*} [ht!]%
       \centering
       \includegraphics [width=0.6\textwidth]{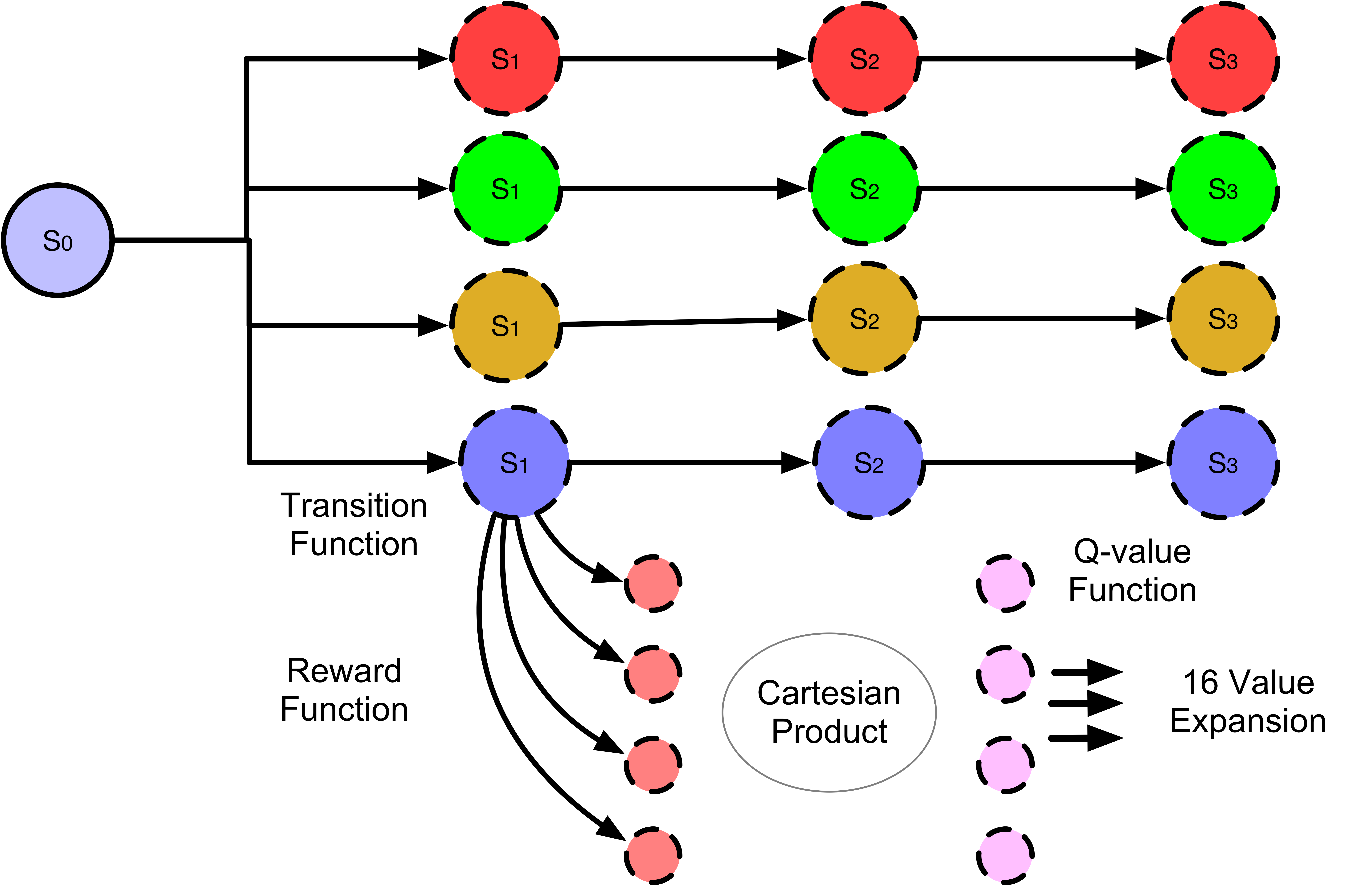}
       \caption{An illustration of the rollout result with $N=4,H_{max}=3, P=4$. For each state of the trajectory, there are 16 candidate targets.}
       \label{fig:Rollout}
\vspace*{-.1in}
\end{figure*}

\subsection{Rollout Details}
 
We employ the identical method of target candidates computation as STEVE, except we image a rollout with an ensemble of probabilistic models. At first, we bind the parameters of transition model ($\zeta_{s,i}$) to the termination model ($\zeta_{d,i}$). That is, we numerate the combination of three integers $\{i,j,k | i,j,k \in  [1,N]\}$, which gives us an ensemble of $N^3$ parameters $\{\zeta_{r,j}, \zeta_{s,i}, \zeta_{d,i}\, \phi'_{k}\}$. The actual sampling process goes like this: For each $H \in  [0, H_{max}]$, we first use the transition model ($\zeta_{s,i}$) and the termination model ($\zeta_{d,i}$) to image a state-action sequence $\{s_{t+1}, \tilde{a}_{t+1}, \tilde{s}_{t+2}, \tilde{a}_{t+2}, ..., \tilde{s}_{t+H+1}, \tilde{a}_{t+H+1}\}$; Based on the state-action sequence, we use reward function ($\zeta_{r,j}$) to estimate the rewards  ($\hat{r}_{t'} = \hat{f}_{r, \zeta_{r,j}} (\tilde{s}_{t'}, \tilde{a}_{t'}, \tilde{s}_{t'+1})$) and the value function ($\phi'_k$) to predict the value of the last state ($\hat{Q}_{\phi'_k} (\tilde{s}_{t+H+1}, \tilde{a}_{t+H+1})$) (fig.~\ref{fig:Rollout}). In total we predict $N^3 (H_{max} + 1)$ combination of rewards and value functions in both RAVE and STEVE.

\section{Details for NeurIPS 2019: Learn to Move Challenge}
The \textbf{Learn to Move} challenge is the third competition of the series in NeurIPS 2019, requiring the participants to train a controller for 3D human model to follow the input velocity commands. The agent has to reach two target points sequentially in 2500 frames. At each step, the agent has to follow an input velocity command that depends on its relative position with the target point. Compared with last year's task, there are three challenges: real-time velocity commands, which varies according to the position of the agent; unknown destination at any angle, sometimes at the back of the agent; minimum effort required during the locomotion.

Following the 1st place's solution of last year, we divide our solution into two stages: learning a usual human-like walking gait and learning to follow the input velocity commands. 

\subsection{Stage 1: Sensible Walking Gait}
We believe that the human-like gait is more proper and flexible for real-time velocity targets. However, the agent tends to learn curious walking gaits such as jumping or staggering, when simply setting its target to a lower speed walking. The authors of \cite{DBLP:journals/corr/abs-1902-02441} proposed using curriculum learning to learn a flexible walking gait. The first aim is to run very fast , because the agent has limited potential gaits to move in a high speed. Then it learns to walk at lower speeds gradually, still keeping a human-like gait that two leg moves forward alternately. 

\subsection{Stage 2: Following targets}

\textbf{Distilling}. After attaining a usual gait walking like the human, the main problem is to distill a policy network with velocity targets from the current policy network. As the new network requires input of real-time targets, which has not been considered in the previous training stage, the distilled network perform worse in the new task. To solve the issue, we improve the robustness of the distilled policy by adding noise during the distilling process. Suppose we have collected a dataset $D (x,y)$ from the old policy, and want to fit a new policy function, $f (x,t)=y$, where t is the velocity target. We replace $t$ with an uniform noise in the training process, to make the new function robust with any unseen targets.

\textbf{Low-energy locomotion}. A main challenge in this competition is to finish the target with low energy. We found that adding a large penalty on muscles will degrade the flexibility and robustness of the agent. One explanation is that agent fails to explore efficient gaits due to the limitation of muscle penalty. Our solution is training an agent without muscle penalty to facilitate the exploration of action spaces at first, then add the muscle penalty to the reward to reduce the energy used.

\begin{figure} [!ht]
\centering

\subfigure [Flexible and efficient starting gait]
{
    \begin{minipage} [t]{0.6\textwidth}
    \centering
    \includegraphics [width=7cm,height=2cm]{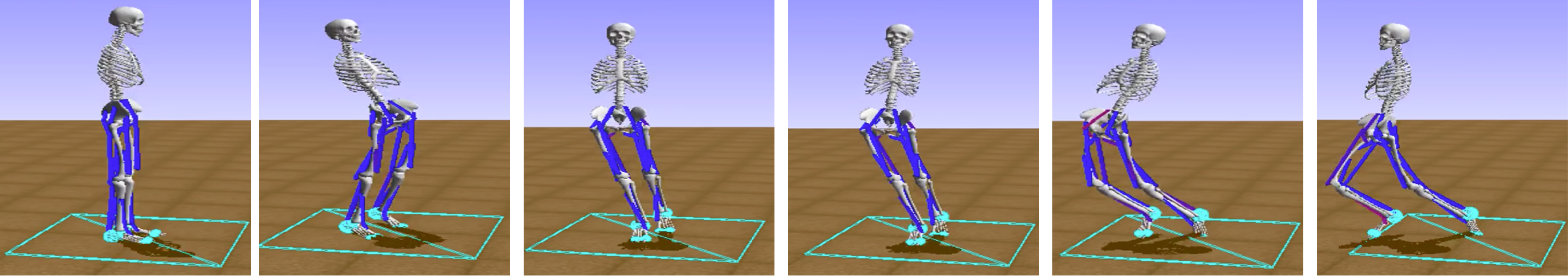}
    \label{fig:track2_final}
    \end{minipage}
}
\subfigure [Velocity gap]
{
    \begin{minipage} [t]{0.3\textwidth}
    \centering
    \includegraphics [width=4cm,height=2cm]{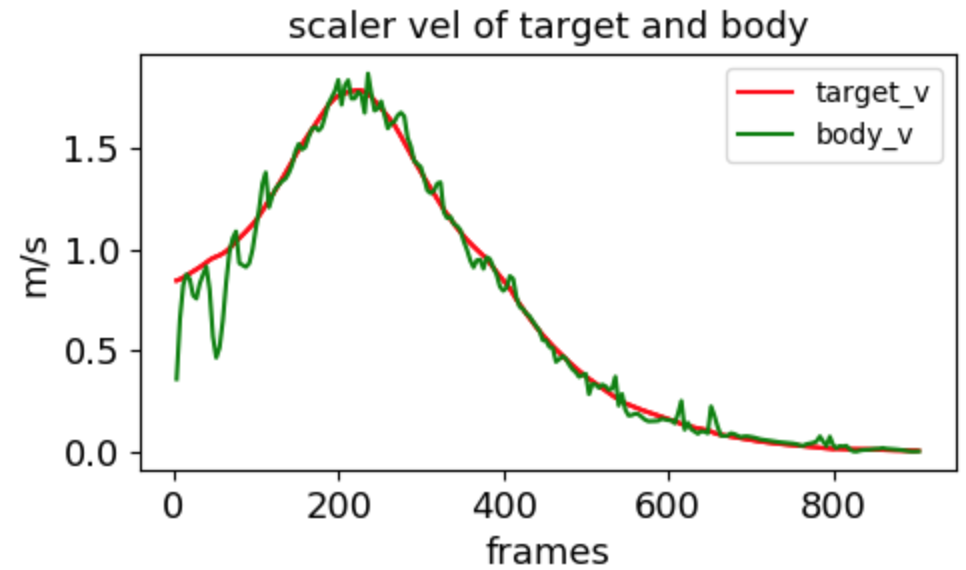}
    \label{fig:target_vel}
    \end{minipage}
}

\caption{Submitted agent. (a) The agent has to turn back to reach the target. (b) The velocity difference between the velocity targets and the current velocity.} 
\vspace*{-.2in}
\label{fig:final-gait}
\end{figure}

\subsection{Note on robustness}
A main concern of the learned agent is its robustness, because the agent gets 500 scores as a bonus for completing the task. To address the issue of falling, previous works\cite{DBLP:journals/corr/abs-1902-02441, DBLP:journals/corr/abs-1712-08987} relies on an ensemble of Q-value functions to evaluate the future returns of candidate actions. However, the estimated Q-values suffer from high bias and variance as the agent has no information about the following velocity commands, which is critical for reward computation.  

We tried using an ensemble of Q-value functions over DDPG, but the learned agent with ensemble action prediction was prone to fall, at the probability of 15\%. When training the agent with RAVE, the falling rate dropped significantly. We evaluate the submitted agent by running 5000 episodes locally. The agent walks to the target points in 2500 frames at each episode, and the mean score is 1489.54, with the probability 1.3\% of falling(see fig.\ref{fig:final-gait} for more information about the learned agent.)

\section{Rewards for Osim-based experiments}
\textbf{LearnToRun} \qquad $r = max(d_l,d_r) + 0.5$

Here $d_l$, $d_r$ is the moving distance On X-axis of the left leg and right leg. \\

\section{Hyper-parameters for Training}
We list all the hyper-parameters used in our Mujoco and Roboschool experiments in table \ref{Tab:Hyperparameters}. For osim tasks, we used the same hyper-parameters of the last year's winning solution\cite{DBLP:journals/corr/abs-1902-02441}.
\begin{table} [h!]
\begin{center} 
  \caption{Table of hyper-parameters for Mujoco and Roboschool experiments}
  \label{Tab:Hyperparameters}
  \begin{tabular}{m{4cm}<{\centering}|m{2.5cm}<{\centering}|m{6.5cm}<{\centering}}
    \hline
    Hyper-parameter & Value & Description \\ 
    \hline
   	$B$ & 512 & Batch size for training the RL, and also the dynamics model \\
   	\hline
   	$N_{rpm}$ & 1e6 & Size of the replay buffer storing the transitions\\
   	\hline
   	$lr_\pi$ & 3e-4 & Learning rate of the training policy\\
   	\hline
   	$lr_Q$ & 3e-4 & Learning rate of the Q-value function \\ 
    \hline
   	$lr_D$ & 3e-4 & Learning rate of the dynamics model \\ 
    \hline
   	$\epsilon$ & 0.05 & Probability of adding a Gaussian noise to the action for exploration\\ 
    \hline
   	$H_{max}$ & 3 & Maximum horizon length for value expansion\\ 
    \hline
   	$N$ & 4 & Ensemble size of the value function and environment models\\
    \hline
   	$F$ & 10000 & Number of collected frames to pretrain the dynamics model before training the policy\\
    \hline
    $Z$ & 1 & Scaling factor for the prediction error \\
    \hline
   	$alpha$ & 1.5 & Confidence lower bound in equation.\ref{RAVE_Equation}\\
    \hline
  \end{tabular}
\end{center}
\end{table}

\end{document}